\documentclass[runningheads]{llncs}

\usepackage{eccv}

\usepackage{eccvabbrv}

\usepackage{graphicx}
\usepackage{booktabs}

\usepackage[ruled,linesnumbered]{algorithm2e}
\usepackage{graphicx}
\usepackage{amsmath}
\usepackage{amssymb}
\usepackage{booktabs}
\usepackage{multirow} 
\usepackage{threeparttable}
\usepackage{bbding}

\usepackage[pagebackref,breaklinks,colorlinks,citecolor=eccvblue]{hyperref}

\begin{document}

\title{Collaboration of Teachers for Semi-supervised Object Detection} 

\titlerunning{Collaboration of Teachers for Semi-supervised Object Detection}

\author{Liyu Chen$^*$ \and Huaao Tang$^*$ \and Yi Wen \and Hanting Chen \and Wei Li \and Junchao Liu  \and \\ Jie Hu}
\authorrunning{L. Chen et al.}

\institute{
	Huawei Noah's Ark Lab\\
	\email{\{chenliyu11, tanghuaao, wenyi14, chenhanting, wei.lee, liujunchao12, hujie23\}@huawei.com}}

\maketitle
\renewcommand{\thefootnote}{\fnsymbol{footnote}}
\footnotetext[1]{These authors contributed equally.}

\begin{abstract}
	Recent semi-supervised object detection (SSOD) has achieved remarkable progress by leveraging unlabeled data for training. Mainstream SSOD methods rely on Consistency Regularization methods and Exponential Moving Average (EMA), which form a cyclic data flow. However, the EMA updating training approach leads to weight coupling between the teacher and student models. This coupling in a cyclic data flow results in a decrease in the utilization of unlabeled data information and the confirmation bias on low-quality or erroneous pseudo-labels. To address these issues, we propose the Collaboration of Teachers Framework (CTF), which consists of multiple pairs of teacher and student models for training. In the learning process of CTF, the Data Performance Consistency Optimization module (DPCO) informs the best pair of teacher models possessing the optimal pseudo-labels during the past training process, and these most reliable pseudo-labels generated by the best performing teacher would guide the other student models. As a consequence, this framework greatly improves the utilization of unlabeled data and prevents the positive feedback cycle of unreliable pseudo-labels. The CTF achieves outstanding results on numerous SSOD datasets, including a 0.71\% mAP improvement on the 10\% annotated COCO dataset and a 0.89\% mAP improvement on the VOC dataset compared to LabelMatch and converges significantly faster. Moreover, the CTF is plug-and-play and can be integrated with other mainstream SSOD methods.
  \keywords{Semi-supervised Learning, Object Detection}
\end{abstract}

\section{Introduction}
\label{sec:intro}

\begin{figure}[h]  %
	\centering
	\includegraphics[scale=0.4]{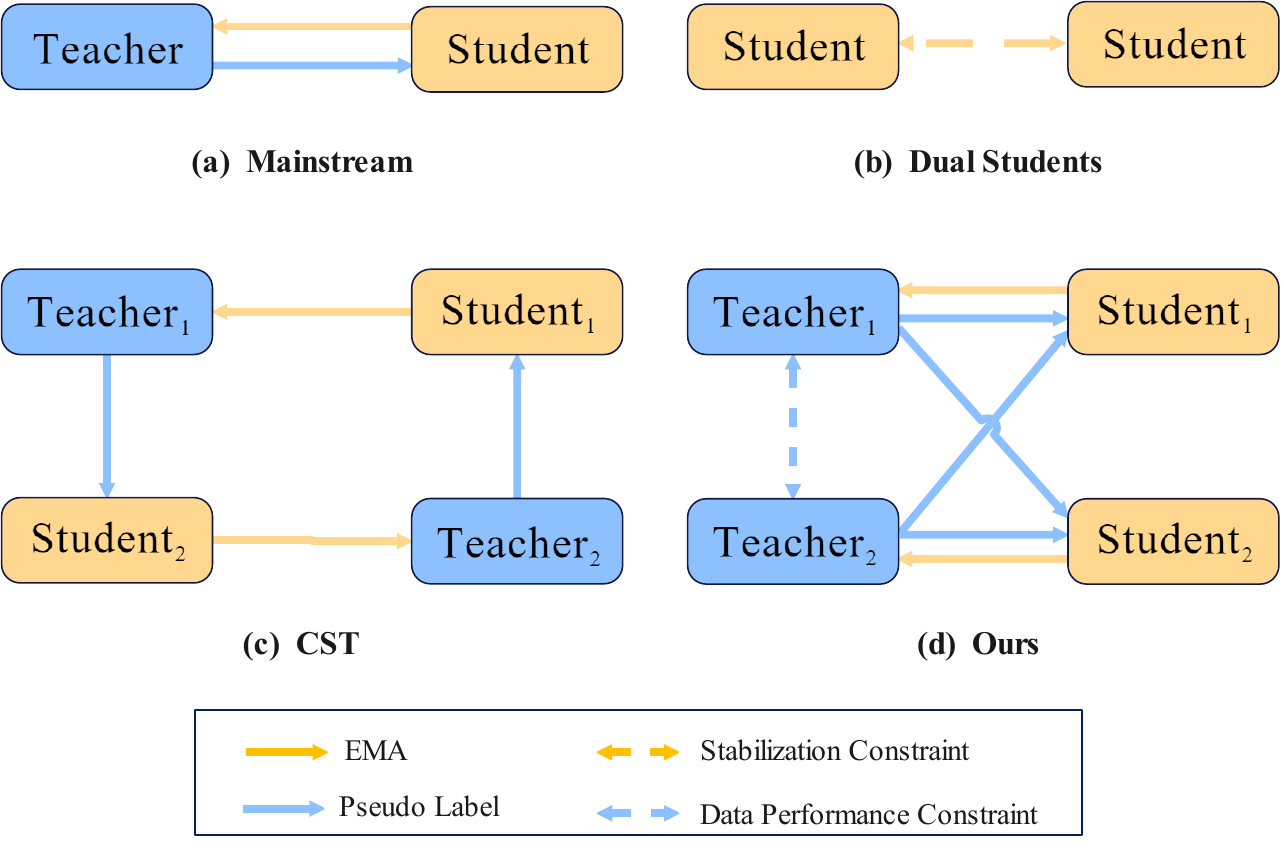}
	\vspace{-0.5em}
	\caption{Figure shows different methods of data flow. (a) The mainstream SSOD method. (b) Differently initialized students with stabilization constraint.\cite{ke2019dual} (c) Two pairs of teacher and student forming a cyclic data flow.\cite{liu2022cycle} (d) Multiple pairs of teacher and student forming a cross data flow.}
	\label{intro_f3}\vspace{-1em}
\end{figure}

In the past few years, the rapid advancements of deep learning triggered soaring developments of plentiful computer vision models, such as those designed for visual classification and detection tasks \cite{girshick2015fast,redmon2016you,lin2017feature,lin2017focal,carion2020end}. These most outstanding detection architectures in the academia are mostly based on supervised training. However, the acquirement of labels, especially the accurate and high-quality ones, is extraordinarily costly and time-consuming. For detection tasks, instance-level labels are even harder to obtain. Such lack of labels gave rise to Self-supervised Learning (SSL) \cite{yang2022survey}, which is a learning paradigm to train the models with a small amount of labeled data combined with a large amount of unlabeled data. In this paper, we mainly focuses on the deficiency in the field of semi-supervised object detection.

Inspired by their preceding field of study SSL, existing mainstream SSOD methods \cite{sohn2020simple,tang2021humble,liu2021unbiased,xu2021end,zhou2021instant,yang2021interactive,mi2022active,kim2022mum,chen2022label,jeong2019consistency,guo2022scale,li2022pseco} include consistency regularization \cite{yang2022survey}, pseudo-labeling \cite{yang2022survey}, a.k.a self-training \cite{yang2022survey}, and Mean Teacher \cite{tarvainen2017mean}. In brief, consistency regularization enforces invariant detection outputs of different views of the same input, whether disturbed by input-level noise, e.g. image augmentation or an adversarial image sample pair, or weight-level and layer-level perturbations of the network; pseudo-labeling produces pseudo-labels for unlabeled instances by considering the high confidence model's predictions of an unlabeled image as its pseudo-labels, which can regularize the training of the model; Mean Teacher is a specific implementation of the pseudo-labeling methods, in which a student model fed by weakly augmented input updates a teacher model fed by strongly augmented images via exponential moving average (EMA) \cite{haynes2012exponential}. These methods together form the commonly used frameworks for SSOD, in which a teacher model pre-trained on labeled data generates pseudo-labels on unlabeled data for a student model to learn and the student model updates the teacher model through EMA.

\begin{figure}[t]
	\centering
	\begin{minipage}[t]{0.45\textwidth}
		\centering
		\includegraphics[width=5.5cm]{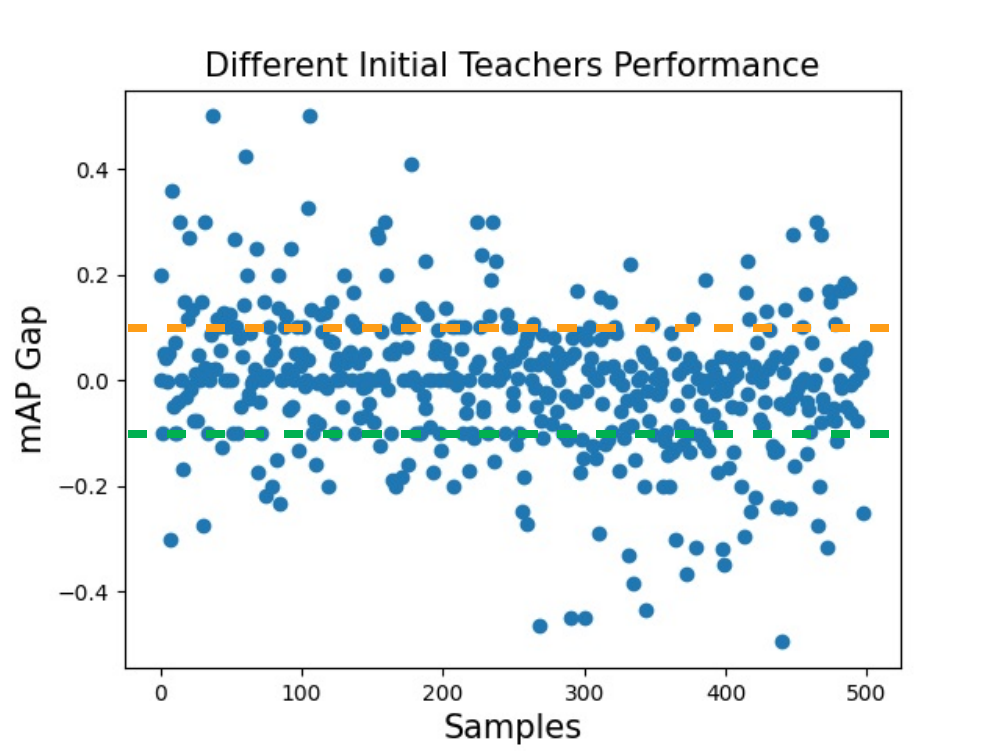}
		\caption{It shows the value of Teacher Model 1 mAP minus Teacher Model 2 mAP for each images. Model 1 consistently outperforms Model 2 significantly above orange line. Conversely, Model 2 exhibits better performance below the green line.}
		\label{map_gap}
	\end{minipage}
	\hspace{5mm}
	\begin{minipage}[t]{0.45\textwidth}
		\centering
		\includegraphics[width=5.5cm]{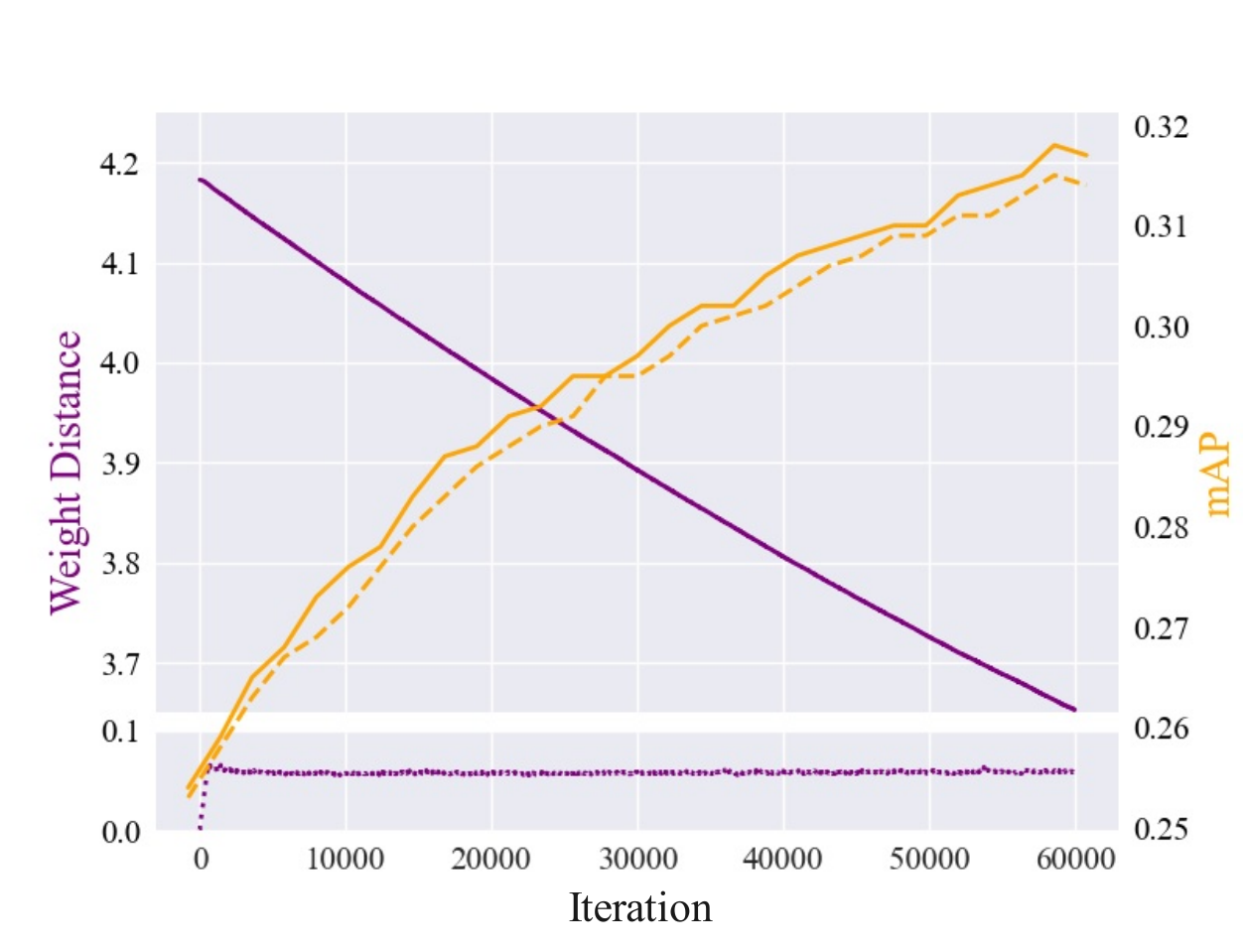}
		\caption{The relationship between model weight distances and performance. The dotted line represents LabelMatch\cite{chen2022label}, while the solid line stands for our approach.}
		\label{intro_f2}
	\end{minipage}
	\vspace{-1em}
\end{figure}

Though these past SSOD methods have achieved excellent results, several fundamental issues overlooked in these SSOD designs hinder the progress of these methods' performances. Firstly, the current SSOD methods generally suffer from the issue of teacher-student over-coupling, where the weights of the teacher and student models become similar during the semi-supervised training process. Consequently, the outputs of the same sample after passing through both models become highly similar, limiting the student model's ability to learn useful information from the teacher model and thereby decreasing the model's performance as shown in the Figure \ref{intro_f2}. Secondly, Figure \ref{intro_f3}.a shows that the confirmation bias arises such that the low-quality or inaccurate pseudo-labels can continuously degrade the learning of the teacher and the student in such a positive feedback training scheme when unreliable information is learned in a circular manner. To tackle the issue above, CST \cite{liu2022cycle}, consisting of two pairs of teacher and student, forms a cycle data flow as shown in Figure \ref{intro_f3}.c, which uses a cyclic update pattern to relax the excessive coupling. However, this information propagation method still involves a positive feedback loop of data flow, resulting in the issue of confirmation bias. The other research dual student \cite{ke2019dual} introduces a stability constraint to address confirmation bias. The framework consists of two students that are updated simultaneously, as shown in Figure \ref{intro_f3}.b, where the more stable student guides the update of the less stable student based on their performance on the current sample. However, using stability constraint to evaluate the performance of models in the semi-supervised paradigm can cause two problems: (1) the stability constraints only reflect the models' ability to predict consistent pseudo-labels under perturbations, which is different from their ability to predict accurate and correct labels; (2) the stability constraint is calculated on single samples with potentially inaccurate pseudo-labels so that it merely considers the comprehensive information learned in the current iteration and lacks statistical significance when evaluating model performance.

In order to resolve the strongly coupling problem and alleviate the weight similarity among models, we introduce a Co-teaching inspired two-stage training architecture called Collaboration of Teacher Framework (CTF) with multiple pairs of teacher-student models trained independently and simultaneously. As shown in Figure \ref{intro_f2}, during the training process, it can be ensured that the models in different pairs have a certain weight distance, which aids the student models in acquiring more knowledge and reaching higher mAP. To tackle the problem of confirmation bias and the flaws of stability constraint, we propose Data Performance Consistency Optimization (DPCO). Through experiments, we discover that although different initialized models exhibit similar mAP on the overall validation set, their performance on individual samples varies as illustrated in Figure \ref{map_gap}. It suggests that the insufficient labeled data will cause different teacher models to converge to different local optima during the supervised pre-training.  Therefore, if we can utilize the pseudo-labels guided by the most reliable teacher model instantly to guide the student model, we can avoid the confirmation bias caused by the circulation of low-quality data when a single teacher model produces unreliable pseudo-labels. Additionally, we believe that the model should perform consistently between labeled data and unlabeled data from the same distribution in a presence of sufficient data. Based on this assumption, we experimentally validated that our proposed DPCO module, which selects reliable teachers based on accumulated sample loss, is more consistent compared to the stability constraints method proposed by the dual student \cite{ke2019dual} and the single-sample method as shown in the Figure \ref{Exp_f2}. The CTF and DPCO module work collaboratively, allowing unsupervised data to be independently trained in differently initialized teacher-student pairs while ensuring efficient and reliable information transfer among pairs. Finally, we are able to obtain a statistically reliable model better at predicting pseudo-labels accurately, both at the classification-level and the localization-level. 

Overall, the main contributions of this work are:

\begin{itemize}
	\item[$\bullet$] We first conduct an analysis of the drawbacks associated with stability constraints and empirically demonstrated the problems, providing a new direction for the evolution of SSOD methods.
	\item[$\bullet$] We propose a Collaboration of Teacher Framework (CTF) to address the coupling issue among teacher-student models, allowing a certain weight distance of different models and different views on samples.
	\item[$\bullet$] We introduce a new constraint for training, the Data Performance Consistency(DPC), which is able to evaluate model performance on unlabeled data without ground-truth (GT) based on labeled data. Cooperating with CTF, DPC can inform which teacher-student pair possesses the most reliable pseudo-labels and guide other pairs in the semi-supervised training.
\end{itemize}

\section{Related Work}

\textbf{Semi-Supervised Object Detection (SSOD).} As introduced aforehand, semi-supervised object detection can be categorized into consistency-based schemes \cite{jeong2019consistency,guo2022scale,li2022pseco} and pseudo-labeling, or self-training, schemes \cite{sohn2020simple,tang2021humble,liu2021unbiased,xu2021end,zhou2021instant,yang2021interactive,mi2022active,kim2022mum,chen2022label}. The latter schemes, generally based on the Mean Teacher \cite{tarvainen2017mean} architecture, are the prevailing schemes for SSOD nowadays. 

As the earliest SSOD work, CSD \cite{jeong2019consistency}, a consistency-based framework, exploits Jensen-Shannon divergence as a measure of left-right-flip consistency to enforce a constant prediction between a pair of horizontally flipped images. A later consistency-based framework PseCo \cite{li2022pseco} enforces the feature-level multi-view scale consistency by aligning the shifted pyramid features of different scale inputs of the same image. Similarly, SED \cite{guo2022scale} also introduces scale consistency by alleviating the noises from the False Negative samples and class-imbalance via self-distilling region proposals and a classification-loss re-weighting mechanism. As the first to apply a teacher-student architecture to SSOD, STAC \cite{sohn2020simple} proposes a simple multi-stage training framework to combine self-training with consistency training. On labeled data, STAC \cite{sohn2020simple} trains a teacher model to generate pseudo-labels on unlabeld data. The high-confidence pseudo-labels from unlabeled data would then be fed back into the network with strong augmentations for model fine-tuning. As a simplification of the above multi-stage training framework, a Mean Teacher \cite{tarvainen2017mean} inspired end-to-end teacher-student framework, in which the student model updates the teacher model with EMA \cite{haynes2012exponential} in an online manner, is proposed and become the foundation of numerous subsequent SSOD works \cite{sohn2020simple,tang2021humble,liu2021unbiased,xu2021end,zhou2021instant,yang2021interactive,mi2022active,kim2022mum,chen2022label}. Despite using EMA to update the teacher model with the student weights, Unbiased Teacher \cite{liu2021unbiased} especially uses Focal Loss \cite{lin2017focal} to remedy the class-imbalance problems in object detection tasks. Humble Teacher \cite{tang2021humble} and Soft Teacher \cite{xu2021end} leverage dense region proposals, soft pseudo-labels and soft-weight to generate more reliable pseudo classes and boxes, respectively. Label Match \cite{chen2022label} argues that soft labels and soft-weight are insufficient for solving label mis-assignment and proposes different filtering threshold calculated from the label distribution for different classes of pseudo-labels to improve pseudo-label quality. De-biased Teacher \cite{wang2023biased} abandons IoU matching and pseudo-labeling to directly generate favorable training proposals for consistency learning. 

Some other methods boost SSOD performance from the data perspective, such as Instant Teaching \cite{zhou2021instant} , MUM \cite{kim2022mum} and Active Teacher \cite{mi2022active} , each making use of Mixup and Mosaic augmentations, a lossless and stronger mix and unmix data augmentation, and different data initializations according to samples' difficulty, information and diversity. Revisiting Class Imbalance \cite{kar2023revisiting} proposes an adaptive thresholding mechanism to screen optimal bounding boxes and a Jitter-agging module to predict accurate localizations. Mix Teacher \cite{liu2023mixteacher} introduces a framework with a mixed scale teacher to boost the generated pseudo-labels and the scale-invariant learning. Some work focus on dense-guided SSOD, including Dense Teacher \cite{li2022dtg} and ARSL \cite{liu2023ambiguity} whereas some focus on data-uncertainty guided approaches \cite{wang2021data}. While the aforementioned SSOD frameworks are designed for two-stage object detectors with anchors, some other works focus on the one-stage \cite{redmon2016you}, anchor-free \cite{tian2022fully} or DETR-based detectors \cite{carion2020end}. For instance, Efficient Teacher \cite{xu2023efficient} designs an SSOD framework for YOLOV5 \cite{ultralytics2021yolov5}. DSL \cite{chen2022dense} and UBv2 \cite{liu2022unbiased} are the first two to adapt semi-supervised training paradigm for an anchor-free detector. Semi-DETR \cite{zhang2023semi} adapts the SSOD framework to a transformer-based object dtector. Though achieving outstanding performance, these past SSOD frameworks are all confined by the limis of EMA inherent in the Mean Teacher based architecture. Nevertheless, these past methods are all biased by the consistency loss, which is defined to depict a detector’s ability to predict constant outputs under perturbations but not the ability to produce accurate label classes or localizations.

\noindent\textbf{Training Schemes with Multiple Teachers.} As the preceding field of study of Semi-Supervised Object Detection, Semi-Supervised Classification (SSC) methods are also divided into consistency regularization methods, pseudo-labeling methods and hybrid methods. Within each category, there exist various architectures that involve multiple teachers or teacher-student pairs to boost the SSC performances. 

Based on Mean Teacher \cite{tarvainen2017mean}, Dual Student \cite{ke2019dual} with two pairs of decoupled teacher-student models is designed with the stability constraint to improve SSC performance by overcoming the confirmation bias deep-rooted in EMA. Deep co-training \cite{blum1998combining}  \cite{qiao2018deep} framework involves a collaborative training of two networks given two different and complementary views of the same input with the help of View Difference Constraint \cite{qiao2018deep}. Tri-training \cite{zhou2005tri} learns three classifiers from three different training sets obtained by utilizing bootstrap sampling \cite{davison1997bootstrap}. Tri-net \cite{dong2018tri}, inspired by Tri-training \cite{han2018co}, adds random noise to the labeled sample via output smearing \cite{breiman2000randomizing} to generate different training sets for learning three initial modules, which predict the pseudo-label for unlabeled data. The predictions of two modules for unlabled instances consistent are considered confident and stable. 

In the field of Co-teaching methods, Co-Teaching \cite{han2018co} involves the training of two DNNs which teach each other mutually for every mini-batch. In every mini-batch, each of the network treats samples with lower losses as useful samples for training and propogates these useful samples to each of its peer network for updates. Since the two networks are initialized differently and thus learn samples differently, biased selection of "clean" data in any mini-batch can be reduced when the two networks mutually decides the actual useful samples based on the sample losses. Co-teaching+ \cite{yu2019does} further introduces the disagreement strategy of decoupling \cite{malach2017decoupling} and keeps only prediction disagreement data after feeding forward all data. Furthermore, Nested+Co-teaching \cite{chen2021boosting} applies a simple compression regularization named Nested Dropout \cite{rippel2014learning} to boost Co-teaching performance in the presence of label noise. Finally, stochastic co-teaching \cite{de2023stochastic} introduces stochasticity to select or reject training instances which can cope with unkonwn distributions of label noise while training neural networks.

\section{Method}
\begin{figure*}[t]  %
	\centering
	\includegraphics[scale=0.7]{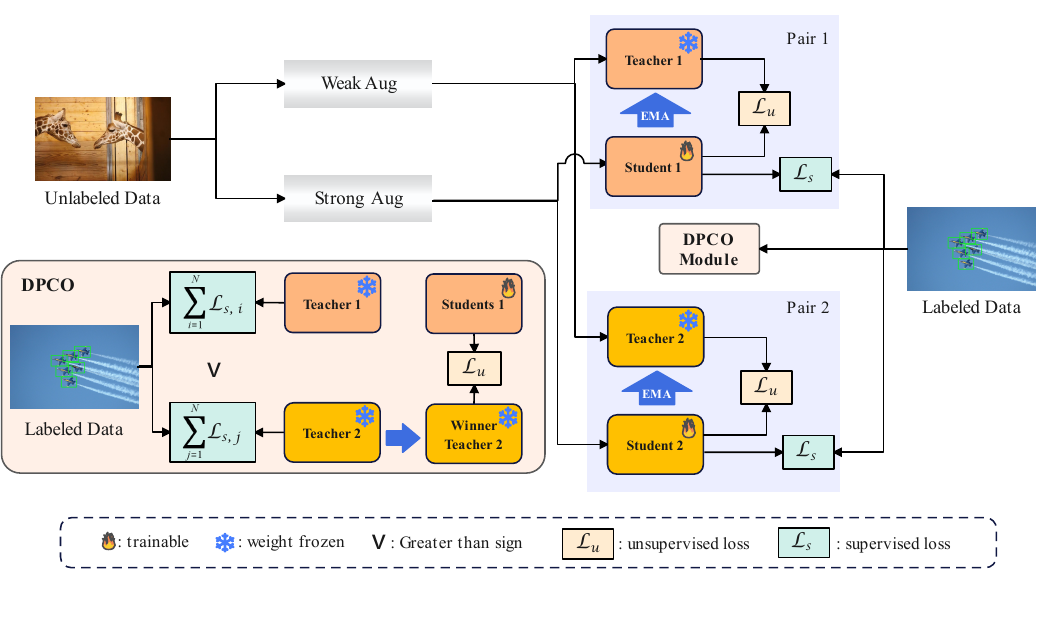}\vspace{-2em}
	\caption{An overview of the Collaboration of Teachers Framework (number of pairs=2). Before training starts, the teacher and student within the same pair have the same weights, while the models between different pairs have different weights. During training, the DPCO module calculates the accumulative labeled loss to select the most reliable pseudo-labels, allowing them to propagate among different pairs.} 
	\vspace{-0.5em}
	\label{f1}
\end{figure*}

We briefly summarize the core idea of Mean Teacher in Sec 3.1. Then, we introduce our Collaboration of Teachers Framework in Sec 3.2. Finally, DPCO and accumulative loss will be elaborated in Sec 3.3. 
\subsection{Overview}
The overall architecture of Collaboration of Teacher Framework is illustrated in Figure \ref{f1}, which involves collaborations of two pairs of decoupled teacher-student models. We adopt the mainstream SSOD paradigm, the Mean Teacher framework, with a pair of detectors to form the teacher-student structure. Within a Mean Teacher framework, the teacher is an exponential moving average of the student detector as shown in the left Figure \ref{intro_f3}. Weakly augmented unlabeled images are fed into the teacher model to generate pseudo-bboxes used for supervising the student detector fed by strongly augmented unlabeled images. Simultaneously, the student detector learns discriminate representation for both classification and regression from labeled images. Given a teacher detector $f_t(\cdot)$ and a student detector $f_s(\cdot)$ that minimize a total loss ${L_{total}}$ which is a weighted sum of the labeled loss ${L_l}$ and the unlabeled loss ${L_u}$:
\begin{equation} \label{eqn}
	{L_l} = \frac{1}{N_i} \sum_i [\mathcal{L}_{cls}(f_s(x_i^l), y_i^l) + \mathcal{L}_{reg}(f_s(x_i^l), y_i^l)] 
\end{equation}
\vspace{-2mm}
\begin{equation} \label{eqn}
	{L_u} = \frac{1}{N_j} \sum_j [\mathcal{L}_{cls}(f_s(x_j^u), \hat{y}_j^u) + \mathcal{L}_{reg}(f_s(x_j^u), \hat{y}_j^u)]
\end{equation} \vspace{-2mm}
\begin{equation} 
	{L_{total}} = {L_l} + \lambda_u{L_u}\label{e1}
\end{equation}
Where $x^l$ and $x^u$ indicates labeled images and unlabeled images respectively, $y^L$ is the ground truth of labeled images, while $\hat{y} = f_t(x^u; W_t)$ is the pseudo-bboxes generated by the teacher model. The teacher parameters $W_t$ is updated by student parameters $W_s$ via EMA:
\begin{equation} \label{eqn}
	W_t = (1-\alpha)W_t + \alpha W_s
\end{equation}

\vspace{-0.5em}
$\lambda_u$ is a weighting parameter to balance the unlabeled loss and the labeled loss. For a fair comparison, we adopted focal loss as the classification loss $\mathcal{L}_{cls}$ and GIoU loss as the regression loss $\mathcal{L}_{reg}$.

However, past approaches have suffered from significant issues. For instance, Figure \ref{intro_f2} illustrates that the excessive coupling of weights between the teacher and student during training, as well as the training paradigm involving regularization consistency and EMA positive feedback loop, continuously deteriorate the model with incorrect pseudo-label information, ultimately leading to confidence bias. These are pressing problems that urgently need to be addressed in the field of SSOD.

\begin{algorithm}[t]
	\scriptsize
	\SetAlgoLined
	\caption{Collaboration of Teachers Framework}
	\label{A1}
	\KwData { $(x^l,y^l),x^u$: labeled and unlabeled data}
	\KwIn {$(f_{t_i},f_{s_i}),i \in \{1, 2, \cdots, n\} $: $n$ pairs of teacher-student initialized by different $seed_i$}
	\KwOut {$f_{t_k}$: best teacher model for inference}
	\textbf{CTF Phase}\\
	\For {$iter = 1$ \rm{\textbf{to}} $max\_iter$}
	{
		\eIf {iter \rm{in "Stage 1"}}
		{
			\textbf{Generate Pseudo Labels}\\
			$\hat{y}^u_i \leftarrow f_{t_i}(x^u), i \in \{1, 2, \cdots, n\} $\\
			\textbf{Compute Total Loss}\\
			using $(x^l,y^l),(x^u,\hat{y}^u_i)$ to train $f_{s_i}, i \in \{1, 2, \cdots, n\}$ with Equation \ref{e1}\\
			\textbf{Compute Accumulative Loss} (no backpropagation)\\
			$L^{acc}_i \leftarrow L^{acc}_i + L_l(f_{t_i}(x^l),y^l), i \in \{1, 2, \cdots, n\}$\\
			\If{iter \rm{at "Stage 1 end"}}
			{
				\textbf{Decide Winner Teacher}\\
				$k \leftarrow \mathop{\arg\min}\limits_{i}L^{acc}_i$\\
				$L^{acc}_i \leftarrow 0, i \in \{1, 2, \cdots, n\}$
			}
		}
		{
			iter is in "Stage 2"\\
			\textbf{Generate Pseudo Labels}\\
			$\hat{y}^u_i \leftarrow f_{t_i}(x^u), i \in \{1, 2, \cdots, n\} $\\
			$\tilde{y}^u_i \leftarrow f_{t_k}(x^u), i \in \{1, 2, \cdots, n\},i \neq k$\\
			\textbf{Compute Total Loss}\\
			using $(x^l,y^l),(x^u,\hat{y}^u_i)$ to train $f_{s_i}, i = k$ with Equation \ref{e1}\\
			using $(x^l,y^l),(x^u,\hat{y}^u_i),(x^u,\tilde{y}^u_i)$ to train $f_{s_i}, i \in \{1, 2, \cdots, n\}, i \neq k$ with Equation \ref{e2}
		}
	}

\end{algorithm}
\subsection{Collaboration of Teachers Framework}
The prevalent SSOD frameworks based on Mean Teacher are flawed by a tight coupling of the weights between the teacher and student models. Hence, the similar weights lead to similar pseudo-labels given the same sample, hindering the student from learning meaningful information in the discrepancy between the teacher's pseudo-labels and student's own predictions. Furthermore, most mainstream SSOD methods follow the training paradigm of a cycle data flow introduced in Mean Teacher as shown in Figure \ref{intro_f3}, which can cause confirmation bias. Specifically, in the presence of inaccurate or wrong pseudo-labels generated by the teacher model, student model supervised by these unreliable information can in return contaminate the teacher model by updating the teacher model through EMA. Such a positive feedback training paradigm leads to a continuous deterioration of the models with the low-quality pseudo-labels in the subsequent training. To remedy the aforementioned issue, we propose the Collaboration of Teacher Framework (CTF) with the DPCO module. There are two main phases in our semi-supervised training: (1) the conventional burn-in phase, in which the detectors is pre-trained by labeled data in a supervised way and (2) the proposed Collaboration of Teacher Framework for semi-supervised training.

\noindent\textbf{Burn-In Phase}: In this phase, the teacher models, as the pseudo-label generators in the subsequent training phases, are pre-trained on labeled data before embarking on semi-supervised training. In order to ensure that the best pseudo-labels are selected from the teachers' predictions, we let each independent teacher-student pair learn different perspectives of the same input samples via different model weight initializations. Hence we shuffle the order of input data N times given N pairs of models in the CTF framework, train the ith($i\in N$) teacher with $seed_i$ of data and then copy the weight to the students:
\begin{equation} \label{eqn}
	W_{t_i} \leftarrow W_{t_i}+\gamma \frac{\partial\left(\mathcal{L}_{\text {l}}\right)}{\partial W_{t_i}};\quad
	W_{s_i} \leftarrow W_{t_i}
\end{equation}
After such initialization, the weights of the teacher and student models within the same pair are identical, while the weights differ across different pairs. After the burn-in phase, these models will be used in the CTF training phase to further enhance their performance.

\noindent\textbf{CTF Training Phase}: In the CTF training phase, we provide different model weights to offer different perspectives on the samples and then employ the DPCO module to filter out the best-quality pseudo-labels, enabling efficient and reliable utilization of unlabeled data for training. As analyzed before, most of the mainstream semi-supervised detection models suffer from the problem of excessive coupling between the teacher and the student models. When the teacher and the student's weights are close to each other, their predictions would also be similar, thereby preventing the student model from learning more informative knowledge from the differences between its own outputs and the teacher's.

Specifically, CTF consists of multiple pairs of teacher-student models with the same structure as shown in Figure \ref{f1}, which are pre-trained under supervised data in the burn-in phase. In the first stage of CTF, each pair is trained for a specific number of iterations following the semi-supervised training paradigm. Since each pair's training is relatively independent, the weight distances between each pair's models gradually increase at the end of the first stage. In this process, the DPCO module accumulates loss on labeled data of each model pair, which is independent of the model back propagation. At the end of the first stage, the DPCO compares the accumulative losses among all pairs to determine the outperforming one. In the second stage of CTF, the winning teacher $f_{t_i}$ will be used to guide the training of the student models from the other pairs $f_{s_j}(j\neq i)$ for the same number of iterations as in the first stage. Given the pseudo-labels of winner teacher $\tilde{y} = f_{t_i}(x_k^u)$, the total loss of other students can be expressed as:
\begin{equation} 
	L_{DPC}=\frac{1}{N_k} \sum_k [\mathcal{L}_{cls}(f_{s_j}(x_k^u), \tilde{y}) + \mathcal{L}_{reg}(f_{s_j}(x_k^u), \tilde{y}]\label{e2}
\end{equation}
\begin{equation} \label{eqn}
	{L_{total}} = {L_l} + \lambda_u{L_u} + \beta{L_{DPC}}
\end{equation}
Where $\beta$ is a hyperparameters and set to 2. These two CTF stages are repeated until all teacher-student models converge. Finally, we select the best-performing teacher model as the representative model for inference phase. Despite the higher computational and memory requirements during the CTF training process, the performance improvement is significant, and the inference latency is as lower as other semi-supervised models. 
The entire training process is described in Algorithm \ref{A1}.

\subsection{Data Performance Consistency Optimizing}
The core problem of semi-supervised object detection algorithms lies in obtaining high-quality pseudo-labels, which are influenced by two factors: the model and the unlabeled data. More specifically, the key issue is how to evaluate the model's performance on the current unlabeled data. Due to the absence of annotations, it is challenging to evaluate the model's performance using metrics such as mAP. Therefore, we propose the Data Performance Consistency to address this issue. In particular, considering the homogeneity between labeled and unlabeled data, when there is a sufficient amount of data, we assume that the model's performance on the unlabeled data should be consistent with its performance on the labeled data. This means that if a teacher model performs better on a certain amount of labeled data, it should also perform better on the unlabeled data.
Given M iteration, if:
\begin{equation} \label{eqn}
	\sum_i^M L_l(f_{t_1}(x_i^l), y^l)>\sum_i^M L_l(f_{t_2}(x_i^l), y^l)
\end{equation}
Assuming we know the ground truth for the unlabeled data, we have:
\begin{equation} \label{eqn}
	\sum_i^M L_l(f_{t_1}(x_i^u), y^u)>\sum_i^M L_l(f_{t_2}(x_i^u), y^u)
\end{equation}

Based on the assumption above, we are able to optimize the training of  CTF in Section 3.2, a framework consisting of N pairs of diverse teacher-student models. On the first stage of the CTF framework, the DPCO module calculates the loss of each teacher model on the same batch of labeled data respectively. After a sufficient number of iterations, we sum up this loss on the labeled data, referred to as accumulative loss. Note that the accumulative loss doesn't backpropate, so it doesn't influence the model updating directly. At the end of the first stage, the accumulative loss will be compared to decide the best model on the unlabeled data used during the accumulation. Then, in the second stage, the same unlabeled data used in the first stage is fed again, and the winning teacher in the first stage will propagates its pseudo-labels to all students.

\section{Experiment}
In this section, we will provide a detailed description of the experimental setup for our study.

\subsection{Experimental Setup}
\textbf{Datasets.} We validated the effectiveness of our method on the MS-COCO dataset and the PASCAL-VOC dataset. For a fair comparison, the experimental settings align with previous works as follows: (1) COCO-PARTIAL: on the train2017 set, we used 5\%, and 10\% of the data as the labeled data, while the remaining portion served as the unlabeled data. We then evaluate the performance on the val2017 subset. (2) VOC-PARTIAL: We employ the VOC2007 dataset as the labeled data and the VOC2012 dataset as the unlabeled data. And we evaluate the model performance on the VOC2007 test set.
By conducting experiments on these datasets with the aforementioned settings, we demonstrate the effectiveness of our approach.

\noindent\textbf{Implementation Details.} The detector model employed is Faster-RCNN [34] with FPN [23] and a ResNet-50 backbone [15]. Our method and model are decoupled, making it applicable to other single-stage or transformer-based detection models. Based on the mmdetection framework, during the training phase, we utilize eight GPUs, adopt the SGD optimization strategy with a momentum of 0.9 and a weight decay of 0.0001. For COCO-PARTIAL, we train the decoupled teacher-student pairs for 160,000 iterations with a learning rate of 0.02 and a total batch size of 64 (32 labeled, 32 unlabeled). For VOC-PARTIAL, we train for 160,000 iterations with a learning rate of 0.0025 and a total batch size of 16 (8 labeled, 8 unlabeled).
\subsection{Experiment Results}
For a fair comparison, we evaluated our method against several previous two-stage SSOD model performance based on Faster R-CNN with ResNet-50 backbone and FPN neck on COCO-PARTIAL and VOC-PARTIAL settings.

\begin{figure}[t]
		\centering
		\begin{minipage}[t]{0.45\textwidth}
		\centering
		\includegraphics[scale=0.35]{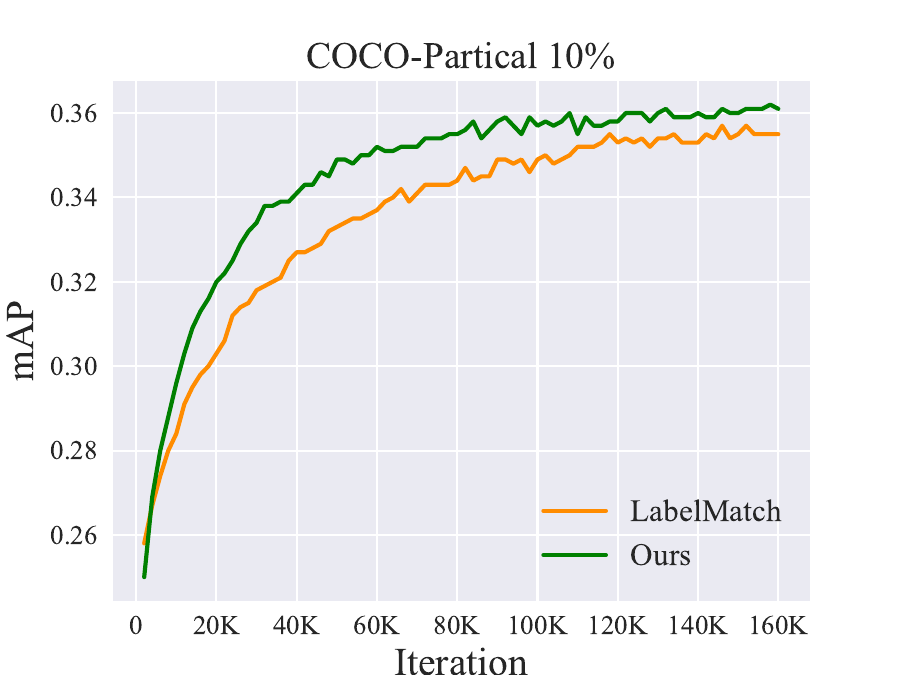}
		\caption{Comparison between the LabelMatch\cite{chen2022label} method and ours during the training process. This experiment was conducted using the 10\% COCO-PARTIAL setting.}
		\label{Exp_f1}
		\end{minipage}
		\hspace{1.5mm}
		\begin{minipage}[t]{0.45\textwidth}
		\centering
		\includegraphics[scale=0.37]{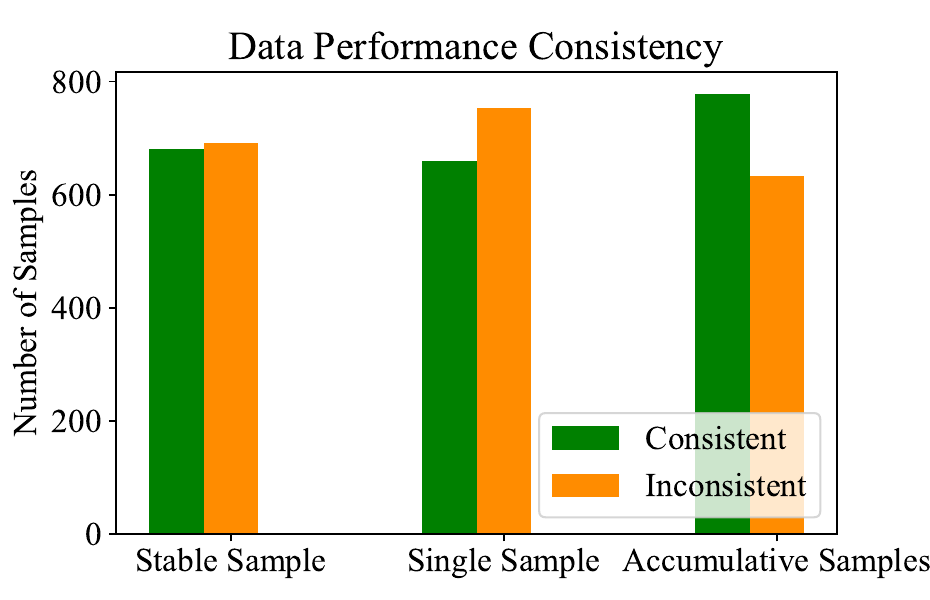}
		\caption{Ablation study of statistical results in different optimizing method for 100 iterations (3200 samples). }
		\label{Exp_f2}
		\end{minipage}

	\end{figure}

\noindent\textbf{COCO-PARTIAL:}
Table \ref{t1} presents the results on 5\% and 10\% labeled COCO datasets, with evaluation metrics using AP50:95. To ensure fairness, all results in the table are evaluated using the two-stage Faster R-CNN detector. It can be observed that compared with training methods using only labeled data, our method shows a significant 
improvement of 15.03\% mAP at 5\%coco setting and 12.34\% mAP at 10\%coco setting, indicating the highly effective utilization of unlabeled data in enhancing model performance. Furthermore, when our approach is integrated as a modular component into the LabelMatch\cite{chen2022label} method, there is a noticeable improvement of 0.8\% mAP at 5\%coco setting and 0.71\% mAP at 10\%coco setting. This demonstrates that our solution, when combined with other state-of-the-art approaches, can achieve higher mAP. When our approach is integrated as a modular component into the Soft Teacher\cite{xu2021end} method, there is an similar improvement of 0.26\% mAP at 5\%coco setting and 1.06\% mAP at 10\%coco setting. Furthermore, our CTF method also demonstrates the advantage of fast convergence during training. Figure \ref{Exp_f1} indicates that our approach achieves the same performance as the LabelMatch\cite{chen2022label} method trained for 160k iterations on 10\% labeled COCO dataset after only 80k iterations, significantly reducing the training time cost.

\begin{table}[t]
	\centering		
	\caption{Experimental results($AP_{50:95}$) of various SSOD methods on \textbf{COCO-PARTIAL} and \textbf{VOC-PARTIAL}. These results based on Faster R-CNN with ResNet-50 backbone and FPN neck. The highest result will be highlighted in bold.}
		\begin{tabular}{@{}l|ccc@{}}
			\toprule
			Methods                  & 5\% COCO & 10\% COCO & VOC       \\ \midrule
			Supervised                  & 18.47    & 23.86  & 42.13        \\ \midrule
			STAC                      & 24.38    & 28.64    & 44.64      \\
			Instant Teaching          & 26.75    & 30.4     & 50.00      \\
			Unbiased Teacher          & 28.27    & 31.5     & 48.69      \\
			Soft Teacher              & 30.74    & 34.04    & 54.60      \\
			ACRST                      & 31.35    & 34.92   & 50.12       \\
			PseCo                       & 32.50     & 36.06 & -         \\
			LabelMatch                & 32.70     & 35.49   & 55.11       \\
			Polishing Teacher           & 32.10     & 35.30 & 52.40          \\
			De-biased Teacher 		 & 32.10     & 35.50    & -       \\
			Soft Teacher + CTF(Ours)\; &\; 31.00\textbf{(+0.26)}         & \;35.10\textbf{(+1.06)}  & \;55.60\textbf{(+1.00)}\\
			LabelMatch +\; CTF(Ours)\; &\; 33.50\textbf{(+0.80)}  & \;36.20\textbf{(+0.71)} & \;56.00\textbf{(+0.89)} \\ \bottomrule
		\end{tabular}
		
		\label{t1}
	\end{table}
	
	\noindent\textbf{VOC-PARTIAL:}
	In addition to validating our approach on the COCO dataset, we also deployed our method on the PASCAL-VOC dataset. For training, we utilized the complete annotated data from VOC2017 and used VOC2012 as unlabeled data. As shown in Table \ref{t1}, the experimental results demonstrate significant improvements achieved by our method. When adapting our approach to LabelMatch\cite{chen2022label}, there is a gain of 0.89\% mAP. Similarly, when adapting our approach to Soft Teacher\cite{xu2021end}, there is a gain of 1.00\% mAP. This indicates the efficient utilization of unsupervised data to enhance model performance across different datasets.

	\subsection{Ablation Study}
	In this section, we will elaborate on the effectiveness of two key designs in our method, namely the Collaboration of Teacher Framework (CTF) and the Data Performance Consistency Optimizing (DPCO) Module. Then, other experimental results will be used to explain specific settings within the framework.

	\noindent\textbf{Collaboration of Teacher Framework.} The ablation experiment performance of each module is shown in Table \ref{t4}. Using CTF alone means that the information transfer between different pairs is random. Although this approach will introduce noise, the training method of cross data flow avoids the negative impact of positive feedback information propagation, resulting in improvements compared to the baseline.

	\noindent\textbf{DPCO Module.} As a key module that guarantees the success of the CTF, DPCO involves the selection of the best performing model among all the teacher-student pairs. After introducing the DPCO module, CTF gains the ability to select the reliable pseudo-labels as supervision signals, filtering out relatively low-quality ones, thereby preventing model from confirmation bias and further enhancing the model's performance as shown in Table \ref{t4}. 
	
	Since there aren't ground-truths for the unlabeled images in the semi-supervised paradigm, it is hard to evaluate the actual detection performances of any model in the teacher-student pairs. Hence, as a replacement of those performance metrics whose calculations involve ground-truths, we construct and assess three evaluation metrics to estimate the detection ability of the models in an absence of data annotations. Figure \ref{Exp_f2} illustrates the effectiveness of the three estimator evaluation metrics, in which Stable Sample means evaluating model following the stability constraint, Single Sample represents estimating teacher performance using the loss of a single labeled data sample, and Accumulative Sample represents the results obtained by the DPCO module using the accumulative loss. `Consistent' indicates the number of samples where the accurate judgments made using the aforementioned method are the same as those made using the ground truth of the unlabeled data. To be more specific, the more 'consistent' samples classified by the estimator metric, the better the estimator metric at evaluating model's detection ability. Otherwise, those results of discrepancy are categorized as `inconsistent'. Note that, since we have the privilege to access ground-truths of the unlabeled data, the evaluation of the effectiveness of each of the estimator metrics involves comparisons with the ground-truths of the data which remain unlabeled while training with CTF. In other words, ground-truths of unlabeled data are used only in the ablations of DPCO. As shown in the Figure \ref{Exp_f2}, it is unreasonable to use the stability consistency and the single loss for accuracy evaluation, while accumulative loss is better at selecting the model with higher accuracy.
	
	\begin{minipage}[!t]{\textwidth}
		
		\begin{minipage}[t]{0.4\textwidth}
			\makeatletter\def\@captype{table}
			\caption{Ablation of CTF and DPCO module on 10\% \textbf{COCO-PARTIAL}. To improve computational efficiency, the number of training iterations for this experiment was set to 300k, and the total batch size was 8. }
			\begin{tabular}{@{}ccll@{}}
				\toprule
				CTF & DPCO module & $AP_{50:95}$        &  \\ \midrule
				\XSolidBrush           & \XSolidBrush            & 34.50        &  \\
				\CheckmarkBold        & \XSolidBrush            & 34.70\textbf{(+0.2)} &  \\
				\CheckmarkBold        & \CheckmarkBold           & 34.80\textbf{(+0.3)} &  \\ \bottomrule
			\end{tabular}
			\label{t4}
		\end{minipage}
		\hspace{9mm}
		\begin{minipage}[t]{0.4\textwidth}
			\makeatletter\def\@captype{table}
			\caption{Ablation study of model performance with accumulative label loss during different training iterations. The AP represents the mean of the last ten results after convergence.}
			\vspace{1em}
			\begin{tabular}{@{}lc@{}}
				\toprule
				Accumulative Loss & $AP_{50:95}$ (mean) \\ \midrule
				25 Iteration      & 33.75               \\
				100 Iteration     & \textbf{33.81}                  \\
				400 Iteration     & \textbf{33.81}                  \\ \bottomrule
			\end{tabular}
			\label{t6}
		\end{minipage}
		\vspace{2em}
	\end{minipage}
	
	Furthermore, we demonstrate that using either the teacher or student model as the representative for evaluating pair accuracy yields similar results by experiments. Under identical conditions, the difference between two strategies is less than 0.1 mAP, and the final model performance tends to be similar as well. Since the teacher model is typically regarded as the mentor in SSOD framework, we ultimately chose the teacher model to calculate the accumulative loss.
	
	Lastly, a crucial aspect of DPCO is the choice of iteration for calculating the accumulative labeled loss. We used 100 iterations as the standard and vary the number by 1/4 and 4 times. As shown in Table \ref{t6}, the model's performance is comparable between 100 and 400 iterations. Therefore, we adopt 100 iterations as our experimental setting.
	
	\vspace{1em}

\section{Conclusion}
In this paper, we meticulously analyze and experimentally identify the existing issues in the field of SSOD. The severe coupling issue intrinsic in the mainstream SSOD training approach based on regularization consistency and EMA, which forms a data positive feedback loop, causing conformation bias when inaccurate pseudo-labels appear. To address the issue of weight coupling between teacher and student models, the CST\cite{liu2022cycle} method proposes a method of cyclically transferring information between two pairs of teacher and student models. However, it still faces the problem of confirmation bias caused by constant flow of low-quality pseudo-labels within this cycle. Furthermore, existing solution of the confirmation bias, dual student\cite{ke2019dual}, can only guarantee the robustness of the model to different augmentations of the same data, but not the accuracy of the model-generated pseudo-labels. To tackle these problems, we propose the CTF with the DPCO module, by which the unlabeled data flow across multiple pairs of teacher-student models to avoid the positive feedback training paradigm and the statistically more reliable teacher model with precise pseudo-labels are selected to ensure better training of the student models.


\bibliographystyle{splncs04}
\bibliography{main}


\clearpage
\newpage

\setcounter{section}{0}
\renewcommand\thesection{\Alph{section}}
\section{Analysis of Loss Accumulation Methods}
\begin{figure}[ht]  %
	\centering
	\includegraphics[scale=0.5]{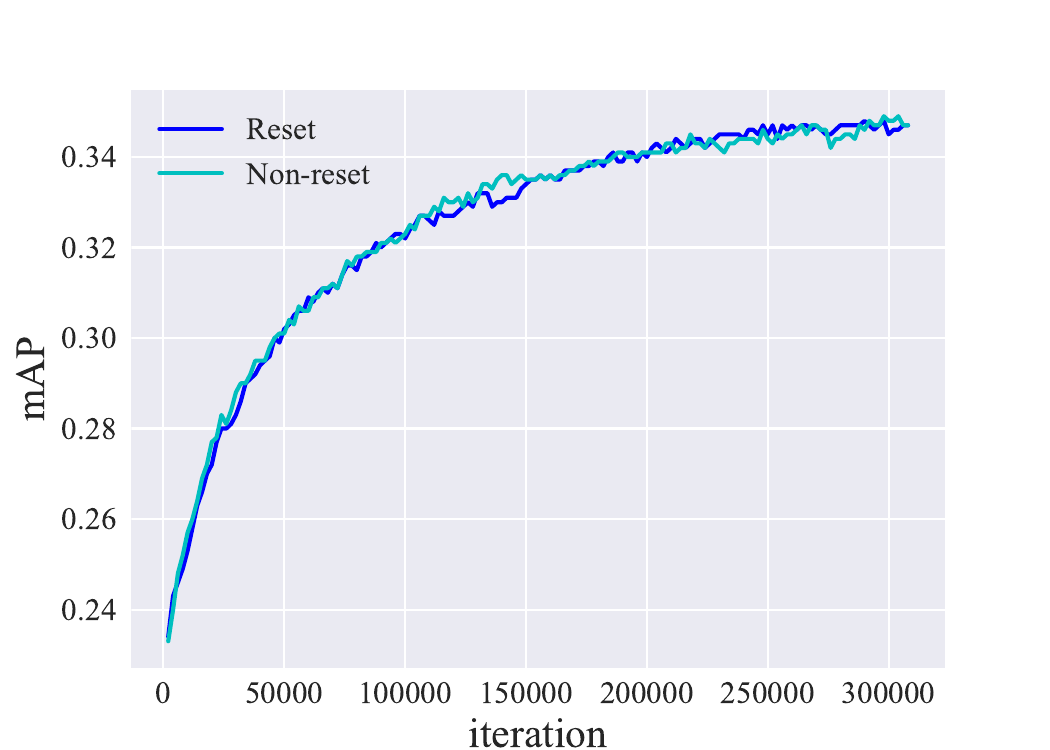}
	\caption{Performance comparison between the reset and non-reset of accumulative loss during the training phase on COCO-Partial settings with 10\% labeled data.}
	\label{s0}
\end{figure}

We conduct and analyze another relevant ablation experiments regarding the label loss accumulation method incorporated in the DPCO module as shown in Figure \ref{s0}. As a reiteration, the DPCO module is to select the best teacher at the end of the first stage of CTF so that the module requires the accumulated loss on the labeled data in the first stage of CTF. We conduct experiments and analysis on the two ways to enable the DPCO module: (1) resetting the accumulated loss after DPCO module have selected the best performing model at the end of the first training stage of CTF and (2) not resetting and continuing loss accumulation based on the previous training stages of CTF. The first way to enable DPCO module by resetting accumulated loss implies that the model's performance on the next round of data in CTF is independent of the performance in the previous rounds, and thus the historical losses from other rounds of training are not considered when selecting the best performing model in the current round. On the contrary, the second way to not reset the accumulated loss concerns about the historical records in other rounds of training. As shown in Figure \ref{s0}, during the 300k training process, the best teacher under both accumulation methods exhibits similar performance. Additionally, from the statistical results, the mean performance over the last ten iterations is nearly the same. Therefore, either method can be used in practice, and we by default reset the accumulative loss in this paper.

\section{Qualitative Results and Analysis}

\begin{figure}[htbp]  %
	\centering
	\includegraphics[scale=0.15]{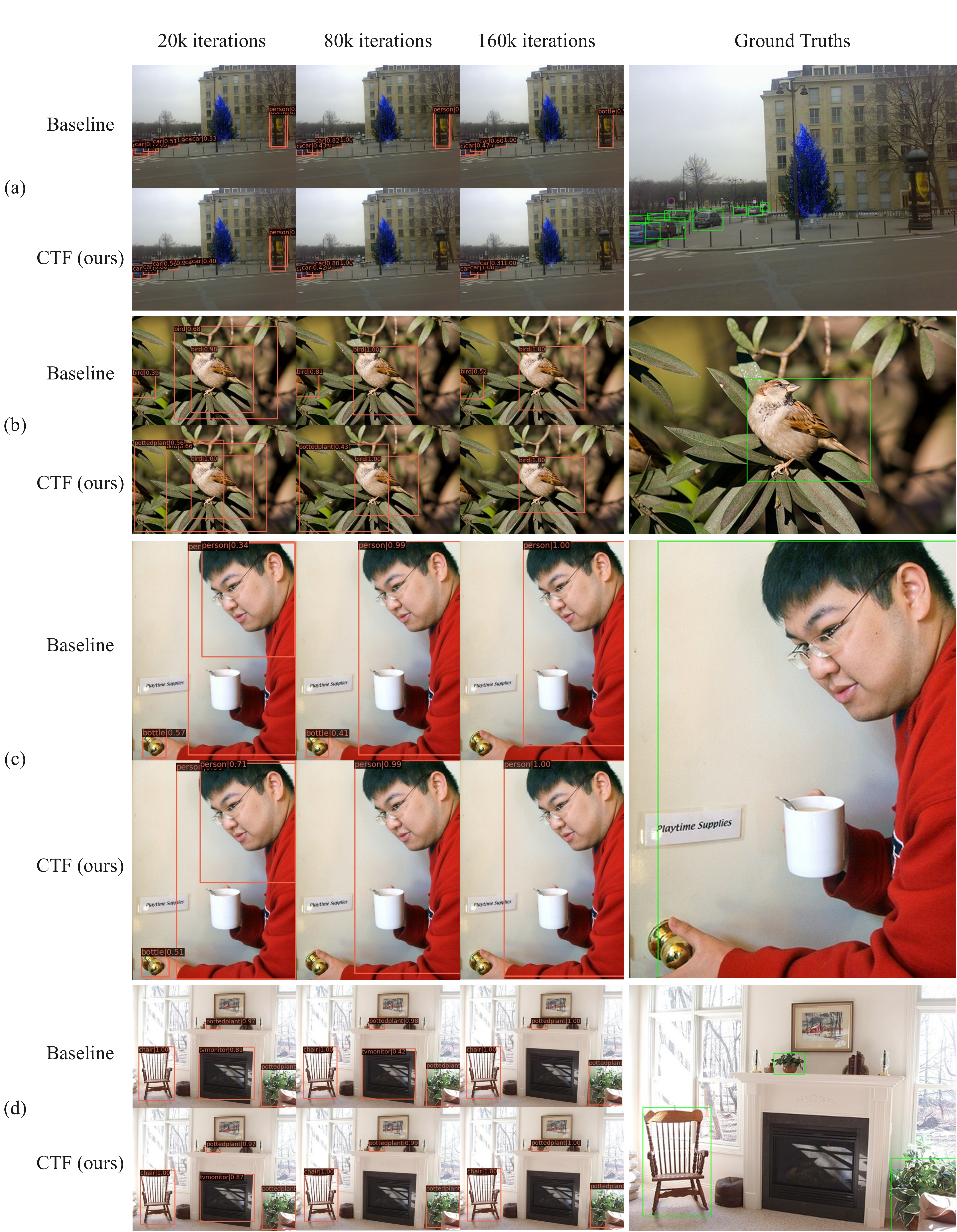}
	\caption{Comparison on confirmation bias (a\&b) and speed of convergence (c\&d) with baseline. We visualize the pseudo labels (red boxes) generated by the baseline SSOD method and CTF (our method) at iterations 20k (close to the start of the training), 80k (in the middle of the training) and 160k (at the end of the training). The ground truths are visualized in green boxes in the rightmost figures, respectively. }
	\label{s3}
\end{figure}

\subsection{CTF Relieves Confirmation Bias}
As demonstrated in earlier sections, CTF is capable of relieving confirmation bias in the presence of noisy and inaccurate pseudo labels. As plotted in Figure \ref{s3} (a), a mis-labeled bottle by the baseline SSOD method never loops out of the positive feedback cycle of the deeply coupled teacher-student network. However, the same inaccurate pseudo label is filtered out by CTF at or before 80k iterations with the help of the cross data flow among pairs of teacher-student models and the effective model selections of the DPCO module. Similarly, in Figure \ref{s3} (b), the baseline method mis-labels a `bird' on the left which is never screened from the beginning till the end of the training. On the contrary, CTF mis-labels a `potted plant' at the beginning of the training but effectively filters the inaccurate pseudo label out at the end. Overall, compared with the baseline SSOD method, our method CTF is able to get rid of unreliable pseudo labels generated during the training process with the aid of multiple pairs and the DPCO module, offering better prediction results at the end of the training.

\subsection{Fast Convergence}
Besides better detection performance, the analysis in the earlier sections also proves that CTF converges faster than the traditional SSOD methods, since the DPCO module selects best performing models much more efficiently and effectively throughout the training process. The visualizations further demonstrate such fast convergence. As shown in \ref{s3} (c), CTF can get rid of the mis-labeled `bottle' at the left bottom corner at or before 80k iterations, much faster than the baseline method. Similarly, Figure \ref{s3} (d) manifests that CTF screens out the inaccurate `tv monitor' pseudo label at or before 80k iterations and the baseline method can't. These phenomenon proves the effectiveness of DPCO, which is used repetitively at the end of the first training stage of CTF to select the best performing model based on the models' actual detection ability. Therefore, CTF is granted with faster selection of best model possessing most accurate pseudo labels and hence faster convergence.

\section{Experiment and Training settings}
\subsection{Dataset}
\noindent\textbf{MS COCO dateset.} The Microsoft Common Objects in Context (MS-COCO) is a large-scale image recognition dataset for object detection, segmentation and captioning tasks containing over 330k images, each annotated with 80 object categories and 5 captions describing the scene. The COCO dataset is widely used in computer vision research and has been used to train and evaluate many state-of-the-art object detection and segmentation models. In our SSOD experiments, we construct COCO-Partial by using 5\% and 10\% of the COCO train2017 as the labeled data while the rest are kept as the unlabeled data. The model performances are evaluated on the val2017 subset. The COCO train2017 subset contains 118k images and the COCO val2017 subset has 5k images.

\noindent\textbf{PASCAL VOC dataset.} The PASCAL Visual Object Classes (VOC) dataset is widely used as a benchmark for object detection, semantic segmentation and classification tasks, consisting of 20 object categories. In our SSOD experiments, we utilize the VOC2007 dataset as the labeled data and the VOC2012 as the unlabeled data and the model performance is evaluated on the VOC2007 testset. The VOC2007 dataset contains a total of 9963 images each with a set of objects out of 20 different classes, making a total of 24640 objects. The VOC2007 dataset is divided into 50\% for training/validation and the other 50\% for testing. The VOC2012 dataset is made of 17,125 pairs of images and bounding boxes of 20 classes.

\subsection{Data Augmentation}
The data augmentations used on labeled data are listed in Table \ref{s5} and those weak and strong augmentations used on unlabeled data are listed in Table \ref{s6}. For a fair comparison, following label matching, we do not implement cutout augmentations on the labeled data in this experiment.

\begin{table}[ht]
	\caption{Data augmentation of labeled data during training process.}
	\label{s5}
	\scalebox{0.72}{
		\begin{tabular}{lll}
			\hline
			\multicolumn{3}{c}{\textbf{Labeled   Data}}                                                                                                                                                                                                                                                                                                      \\ \hline
			Augmentation    & Parameters Setting                                                                                                    & Description                                                                                                                                                                                            \\
			Horizontal Flip & p=0.5                                                                                                                 & None                                                                                                                                                                                                   \\
			Multi-Scale     & ratio=(0.2, 1.8), keep ratio=True                                                                                     & \begin{tabular}[c]{@{}l@{}}The short edge of image is random resized \\ from 0.2l\_short to 1.8l\_short\end{tabular}                                                                                   \\
			Color Jittering & \begin{tabular}[c]{@{}l@{}}(brightness, contrast, saturation, hue) \\      = (0.4, 0.4, 0.4, 0.1), p=0.8\end{tabular} & \begin{tabular}[c]{@{}l@{}}Brightness factor, contrast factor, saturation factor \\ are chosen uniformly from {[}0.6, 1.4{]}, while hue value \\ is chosen uniformly from {[}-0.1, 0.1{]}\end{tabular} \\
			Grayscale       & p=0.2                                                                                                                 & None                                                                                                                                                                                                   \\
			GaussianBlur    & (sigma\_x, sigma\_y) = (0.1, 2.0), p=0.5                                                                              & Apply Gaussian filter with sigma\_x = 0.1 and sigma\_y = 2.0                                                                                                                                           \\ \hline
	\end{tabular}}%
	
\end{table}

\begin{table}[ht]
	\centering
	\caption{Data augmentation of unlabeled data during training process.}
	\label{s6}
	\scalebox{0.65}{
		\begin{tabular}{lll}
			\hline
			\multicolumn{3}{c}{\textbf{Unlabeled Data}}                                                                                                                                                                                                                                                                                                                                                                                                             \\ \hline
			Weak Augmentation   & Parameters Setting                                                                                                                                                                                     & Description                                                                                                                                                                                                              \\
			Horizontal Flip     & p=0.5                                                                                                                                                                                                  & None                                                                                                                                                                                                                     \\
			Multi-Scale         & \begin{tabular}[c]{@{}l@{}}long side=1333, short side=(500, 800), \\      keep ratio=True\end{tabular}                                                                                                 & \begin{tabular}[c]{@{}l@{}}Resize long side of the image to   1333 and short side within \\      500-800, while keeping the height-width ratio unchanged.\end{tabular}                                                   \\ \hline
			
			Strong Augmentation & Parameters Setting                                                                                                                                                                                     & Description                                                                                                                                                                                                              \\
			Horizontal Flip     & p=0.5                                                                                                                                                                                                  & None                                                                                                                                                                                                                     \\
			Multi-Scale         & ratio=(0.5, 1.5), keep ratio=True                                                                                                                                                                      & \begin{tabular}[c]{@{}l@{}}The short edge of image is   random resized from 0.5l\_short \\      to 1.5l\_short\end{tabular}                                                                                              \\
			Color Jittering     & \begin{tabular}[c]{@{}l@{}}(brightness, contrast, saturation, hue) \\      = (0.4, 0.4, 0.4, 0.1), p=0.8\end{tabular}                                                                                  & \begin{tabular}[c]{@{}l@{}}Brightness factor, contrast   factor,  saturation factor are chosen   \\      uniformly from {[}0.6, 1.4{]},  while hue   value is chosen uniformly \\ from {[}-0.1, 0.1{]}\end{tabular} \\
			Grayscale           & p=0.2                                                                                                                                                                                                  & None                                                                                                                                                                                                                     \\
			GaussianBlur        & (sigma\_x, sigma\_y) = (0.1, 2.0), p=0.5                                                                                                                                                               & Apply Gaussian filter with sigma\_x = 0.1 and sigma\_y = 2.0                                                                                                                                                             \\
			Cutout              & \begin{tabular}[c]{@{}l@{}}scale = (0.05, 0.2), ratio = (0.3, 3.3), p=0.7;\\      scale = (0.02, 0.2), ratio = (0.1, 6.0), p=0.5;\\      scale = (0.02, 0.2), ratio = (0.05, 8.0), p=0.3;\end{tabular} & \begin{tabular}[c]{@{}l@{}}Randomly selects a rectangle   region in an image and erases \\      its pixels.\end{tabular}                                                                                                 \\ \hline
	\end{tabular}}
\end{table}

\subsection{Training Setting}
Different training settings are implemented on different datasets and experiments in this study as shown in Table \ref{s2}. For a fast ablation, we ran the training scheme by 40k iterations batch sizes of 8 for both unsupervised and supervised phase of training. Despite those shown in the table, we follow label matching to use the SGD optimizer with a momentum rate 0.9 and weight decay 0.0001 in all experiments.

\begin{table}[htbp]
	\centering
	\caption{Experimental settings on different datasets.}
	\label{s2}
	\begin{tabular}{lccc}
		\hline
		Training Setting                                  & \multicolumn{1}{l}{COCO-PARTIAL} & \multicolumn{1}{l}{VOC-PARTIAL} & \multicolumn{1}{l}{Ablation} \\ \hline
		Batchsize (unsupervised \textbackslash{} supervised) & 32 \textbackslash{} 32             & 8 \textbackslash{} 8              & 8 \textbackslash{} 8           \\
		Number of pairs                                   & 2                                & 2                               & 2                            \\
		Surpervised initialization seed                   & 1\&5                             & 1\&5                            & 1\&5                         \\
		GPU                                               & 8                                & 8                               & 8                            \\
		Learning Rate                                     & 0.01                             & 2.50E-03                        & 0.005                        \\
		Learning rate step                                & -                                & -                               & -                            \\
		Iterations                                        & 160k                             & 160k                            & 40k                          \\
		DPC loss weight                                   & 2                                & 2                               & 2                            \\
		Unsupervised loss weight lambda                   & 2                                & 2                               & 2                            \\
		EMA rate                                          & 0.996                            & 0.996                           & 0.996                        \\
		Multi-scale (strong augmentation)                 & (0.5, 1.5)                       & (0.5, 1.5)                      & (0.5, 1.5)                   \\
		Test score threshold                              & 0.001                            & 0.001                           & 0.001                        \\ \hline
	\end{tabular}
\end{table}

\end{document}